\documentclass[10pt,twocolumn,letterpaper]{article}

\usepackage[pagenumbers]{cvpr} 

\usepackage{graphicx}
\usepackage{amsmath}
\usepackage{amssymb}
\usepackage{booktabs}
\makeatletter
\@namedef{ver@everyshi.sty}{}
\makeatother
\usepackage{tikz}
\usetikzlibrary{shapes,arrows}
\usepackage{circuitikz}
\usepackage{bytefield}
\usetikzlibrary{decorations.pathreplacing}
\usepackage{epsfig}
\usepackage{float}
\usepackage{multirow}
\usepackage{graphicx}

\usepackage[pagebackref,breaklinks,colorlinks]{hyperref}

\usepackage[capitalize]{cleveref}
\crefname{section}{Sec.}{Secs.}
\Crefname{section}{Section}{Sections}
\Crefname{table}{Table}{Tables}
\crefname{table}{Tab.}{Tabs.}

\def\cvprPaperID{*****} 
\def\confName{CVPR}
\def\confYear{2022}

\begin{document}

\title{Soft Adversarial Training Can Retain Natural Accuracy}

\author{Abhijith Sharma\\
The University of British Columbia\\
Canada\\
{\tt\small sharma86@mail.ubc.ca}
\and
Apurva Narayan\\
The University of British Columbia\\
Canada\\
{\tt\small apurva.narayan@ubc.ca}
}
\maketitle

\begin{abstract}
Adversarial training for neural networks has been in the limelight in recent years. 
The advancement in neural network architectures over the last decade has led to significant improvement in their performance. It sparked an interest in their deployment for real-time applications. This process initiated the need to understand the vulnerability of these models to adversarial attacks. It is instrumental in designing models that are robust against adversaries. Recent works have proposed novel techniques to counter the adversaries, most often sacrificing natural accuracy. Most suggest training with an adversarial version of the inputs, constantly moving away from the original distribution.
The focus of our work is to use abstract certification to extract a subset of inputs for (hence we call it 'soft') adversarial training. We propose a training framework that can retain natural accuracy without sacrificing robustness in a constrained setting. Our framework specifically targets moderately critical applications which require a reasonable balance between robustness and accuracy. The results testify to the idea of soft adversarial training for the defense against adversarial attacks. At last, we propose the scope of future work for further improvement of this framework. 
\end{abstract}


\section{Introduction}
\label{sec:introduction}

Deep Neural Networks (DNNs) have been through tremendous advancement in recent times. Unsurprisingly, we have also observed a simultaneous interest among the researchers to adopt deep learning techniques for the models in various applications. Especially, DNNs are being extensively utilised in designing models for computer vision  \cite{voulodimos2018deep}, \cite{khan2018guide} and natural language processing (NLP) \cite{otter2020survey} applications. Although there are many areas where DNNs can be adopted apart from the two stated above, we have settled for conventional machine learning methods due to the lack of interpretability in DNNs. Most of these examples can be attributed to the applications in business contexts, where feature engineering on top of machine learning models \cite{dong2018feature} is adopted to achieve the goals. Even though, the shortcoming of lack of interpretability in DNNs is quite evident, but we cannot take away their credit for powerful behavior. In the past decade, researchers have been quite involved in the process of producing more and more powerful models. It induced an unwavering interest among researchers for its implementation in critical real-time systems. Interestingly, until quite recently, we were ignorant of the robustness aspect of DNN models. It unveiled an unexplored space of DNN's behavior against adversarial attacks. \\
\indent Despite the powerful capabilities of DNNs, the lack of reliability and explainability in the technique restricts it from playing a comprehensive role in a real-time system.  Hence, it became increasingly necessary to understand and explain the behavior of DNNs even in vision or NLP-based applications. Some of the famous works of \cite{goodfellow2014explaining} and \cite{nguyen2015deep} have exposed the vulnerability of neural networks against adversaries. Moreover, \cite{papernot2016transferability} has demonstrated that the attacks are transferable, which implies that it is effortless to design an attack even without the exact knowledge of the model. Even though producing a secure DNN model is an active area of research, authors in \cite{madry2017towards} have shown that the idea is achievable. Another exciting approach towards interpretable artificial intelligence, focusing on developing robust models against attacks, is evident in the work of \cite{gehr2018ai2}. In this work, the authors proposed an idea of abstract approximation of possible perturbation to the inputs in the setting. The theory of abstract interpretation dates back to the 1970s \cite{cousot1977abstract}. Although prominently it was being used as an elegant theoretical framework for automated analyzers, authors in \cite{gehr2018ai2} have presented a new dimension for its utilization in analyzing neural networks.  On similar lines, authors in \cite{singh2019abstract}, and \cite{singh2018fast} have demonstrated certified verification of model with abstract symbolic regions and abstract transformers for operations in the context of neural networks.

\section{Motivation}
\label{sec:two}

Our work combines the idea of adversarial training as proposed in \cite{madry2017towards} and the abstract certification. In this paper, we propose a new training strategy  by amalgamating the conventional adversarial training and abstract interpretation techniques to find a better space of inputs for the adversarial training. 
Typically, in the traditional adversarial training,  we perturb every input in original data space and train the models with the new perturbed adversarial space. 
\begin{figure}[h]
	\hspace{-0.4cm}
	\scalebox{1}{
		\input{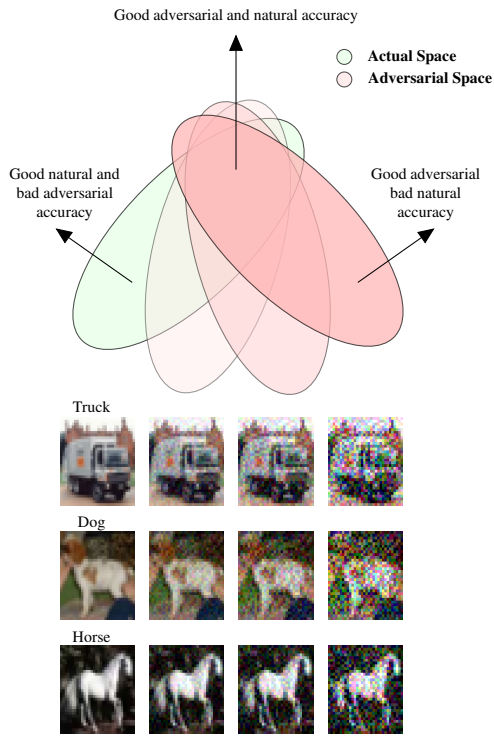}
	}
	\caption{Illustration of Input Space.}
	\label{fig:space}
\end{figure}
However, in this process we tend to sacrifice our natural accuracy as the model is trained on a space different from the original one. The intuition behind our reasoning is as shown in \ref{fig:space}.
In this figure, one can see how we move away from original distribution during adversarial attacks. Hence, in this research we investigate the trade off between severity of adversarial attack and natural accuracy.  
The safety critical applications require strong adversarial training, however, there exist applications that are not overly critical, and we aim for a decent trade-off between accuracy and robustness \cite{kamilaris2018deep}, \cite{you2016image}, \cite{brosnan2004improving} . 
In these scenarios, we do not want to perform adversarial training on the complete set of inputs. In practice, we might observe data points in input space that may not get affected by the adversaries. Hence, in our work, we focus on understanding each of the inputs' behavior through an abstract interpretation-based certification \cite{gehr2018ai2} in a constrained setting. Later, we perform an adversarial training over specific subset of original distribution.

\section{Background}
\label{sec:three}

\subsection{Attacks and Adversarial Training}
There has been extensive work on designing adversarial attacks on neural networks, each having its specificity and specialty \cite{akhtar2018threat}. Subsequently, the motivation to build robust models had led to the design of a defense mechanism for such attacks \cite{xu2020adversarial}. Most of the popular defenses built to date are broadly inspired by the setting as proposed in \cite{madry2017towards} for training against the adversarial attack. In the problem formulation of the defenses, we consider a robust generalization of  the optimization problem as given in \ref{eq:opt}:

\begin{equation}
\min_{x} \rho(\theta) ;  
\rho(\theta) = \mathbf{E}_{(x,y) \sim D} \left[\max_{\delta \in S} \mathcal{L}(\theta,x+\delta,y)\right]
\label{eq:opt}
\end{equation}

The above expression is a saddle point optimization problem with an inner maximization problem aiming for high loss with respect to adversarial example $x+\delta$, where $\delta$ is imperceptible perturbation to the original input x. At the same time, outer minimization problem tries to find the model parameters to reduce the overall expected loss in classification. Here, we incorporate adversarial attacks during the training phase for optimizing the inner maximization problem. One of the famous attacks is Projected Gradient Descent (PGD), which is quite simple to implement and has performed exceptionally well in practice. It is derived from Fast Gradient Sign Method (FGSM) \cite{goodfellow2014explaining}. It is essentially a one-step $l_{\infty}$ bounded adversary, which finds an adversarial example as given in \ref{eq:FGSM}:

\begin{equation}
x+\epsilon.sgn(\nabla_x \mathcal{L}(\theta,x,y))
\label{eq:FGSM}
\end{equation}

\noindent Predominantly, PGD is an iterative version of FGSM with an additional projection utility to ensure the solution lies within the predefined bounds $\epsilon$. This inherent characteristic of PGD makes it more potent than one-step-based attacks like FGSM.  Each iteration in a PGD attack is as given in equation \ref{eq:PGD}:

\begin{equation}
x^{t+1} = \mathcal{P}_{x+S} \; (x^{t}+\alpha.sgn(\nabla_x \mathcal{L}(\theta,x,y)))
\label{eq:PGD}
\end{equation}

\noindent where $\mathcal{P}_{x+S}$ is the projection of the point on the boundary $S$, and $\alpha$ is the step size.
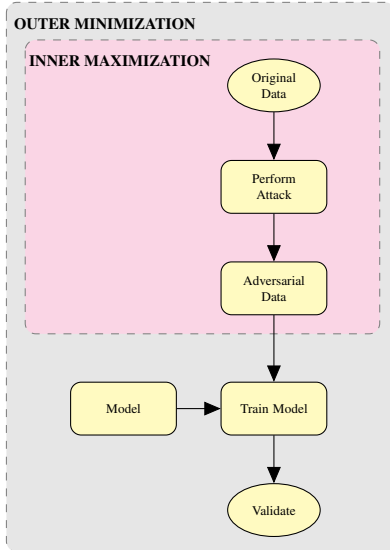
\begin{figure}[h!]
	\hspace{0.2cm}
	\scalebox{1}{
		\centering
\vspace*{-4mm}
\pgfdeclarelayer{background}
\pgfdeclarelayer{foreground}
\pgfsetlayers{background,main,foreground}
\tikzstyle{block} = [rectangle, draw, fill=yellow!30, text centered, rounded corners, minimum height=4em,minimum width=8em,text width=5em]
\tikzstyle{bck} = [rectangle, draw, fill=yellow!30, text centered, rounded corners, minimum height=4em,minimum width=8em,text width=6em]
\tikzstyle{bck1} = [rectangle, draw, fill=yellow!30, text centered, rounded corners, minimum height=4em,minimum width=5em,text width=4em]
\tikzstyle{dia} = [diamond, draw, fill=yellow!30, text centered, rounded corners, minimum height=2em,minimum width=4.5em,text width=4em]
\tikzstyle{diamon} = [diamond, draw, fill=yellow!30, text centered, rounded corners, minimum height=3em,minimum width=6em,text width=5em]
\tikzstyle{ov} = [ellipse, draw, fill=yellow!30, text centered, rounded corners, minimum height=4em,minimum width=7em,text width=4em]	
\tikzstyle{line} = [draw, -latex']
\begin{tikzpicture}[node distance = 3cm, auto,scale=0.5, every node/.style={transform shape}]
\node [ov,align=center,node distance = 2.7cm] (OD) {Original Data};
\node [bck,below of=OD, node distance = 2.7cm] (attack) {Perform Attack};
\node [bck,below of=attack, node distance = 2.7cm, align=center] (ADS) {Adversarial Data};
\node [bck,below of=ADS, node distance = 3.2cm, align=center] (TM) {Train Model};
\node [bck,left of=TM,node distance = 4cm] (model) {Model};
\node [ov,below of = TM, align=center,node distance = 2.7cm] (val) {Validate};
\draw [-triangle 45] (OD) -- (attack);
\draw [-triangle 45] (attack) -- (ADS);
\draw [-triangle 45] (ADS) -- (TM);
\draw [-triangle 45] (TM) -- (val);
\draw [-triangle 45] (model) -- (TM);

\begin{pgfonlayer}{background}
\path (model.west |- OD.north)+(-1.7,1.5) node (a) {};
\path (val.east |- val.south)+(2.0,-0.4) node (b) {};
\path[fill=black!10,rounded corners, draw=black!50, dashed](a) rectangle (b); 
\node [above of=OD, node distance=1.1cm,yshift=0.55cm,xshift=-4.5cm,align=center] (d2) {\large\bf OUTER MINIMIZATION}; 
\end{pgfonlayer} 

\begin{pgfonlayer}{background}
\path (model.west |- OD.north)+(-1.2,0.5) node (a) {};
\path (ADS.east |- ADS.south)+(1.4,-0.5) node (b) {};
\path[fill=magenta!20,rounded corners, draw=black!50, dashed](a) rectangle (b); 
\node [left of=OD, node distance=5.9cm,yshift=0.65cm,xshift=1.8cm,align=center] (d2) {\large \bf INNER MAXIMIZATION}; 
\end{pgfonlayer} 

\end{tikzpicture}
	}
	\caption{Traditional training procedure.}
	\label{fig:adverflow}
\end{figure}
In recent times, the newer upgraded versions outperformed the idea proposed by \cite{madry2017towards}, either in efficiency \cite{shafahi2019adversarial} or speed \cite{wong2020fast}. However, all these settings are similar in that we disturb our original distribution of inputs by generating an adversarial substitute for each of the input space data points.  Figure \ref{fig:adverflow} depicts the design flow of each training epoch in conventional adversarial training. 

\subsection{Abstract Certification}

The response of each input to an adversarial attack might be different. Hence, it is quite reasonable that some inputs might be more vulnerable than the rest. Here, we define an input as vulnerable if it violates a property over outputs called post-condition $\psi$ for a pre-defined input property $\phi$ ($l_{\infty}$ bound in case of brightening attack). The $\phi$, also known as pre-condition, is defined as: 

\begin{equation}
Ball(\emph{x})_{\epsilon}= \{ \acute{\emph{x}} \; \mid \; \| \emph{x}-\acute{\emph{x}} \|_{\infty} < \epsilon \}
\label{eq:precond}
\end{equation}

\noindent where $\emph{x}$ and $\acute{\emph{x}}$ are actual and perturbed input, respectively. The $\psi$ is defined as:

\begin{equation}
\forall \; \emph{j} \in [0,K], \quad o_{k} \geq o_{\emph{j}}
\label{eq:postcond}
\end{equation}

\noindent where $K$ is the number of classes, $k$ is the actual label of the input being verified, and $o_{\emph{j}}$ is the ${\emph{j}}^{th}$ element of the output vector. Even though authors in \cite{madry2017towards} have shown the presence of a local maximum of loss value in a PGD attack, one cannot guarantee the absence of an adversary. In fact, it is infeasible to enumerate all possible adversaries of an input. For example, in the MNIST data set, we have 28 $\times$ 28 = 784 total pixels for each image. Even if we consider the binary perturbation for each pixel (0 for black, 1 for white), we will end up enumerating $2^{784}$ total perturbed version of a single image.  Hence, abstract interpretation of neural networks helps overcome the shortcoming of working with a finite but significant number of images.

In abstract certification,  the output is formulated using abstract transformation of symbolic regions, unlike concrete transformers, where operations are performed over concrete vectors. 
\begin{figure*}[h]
 \center
  \includegraphics[height=31.5cm, width=13cm, angle=-90, scale=0.5]{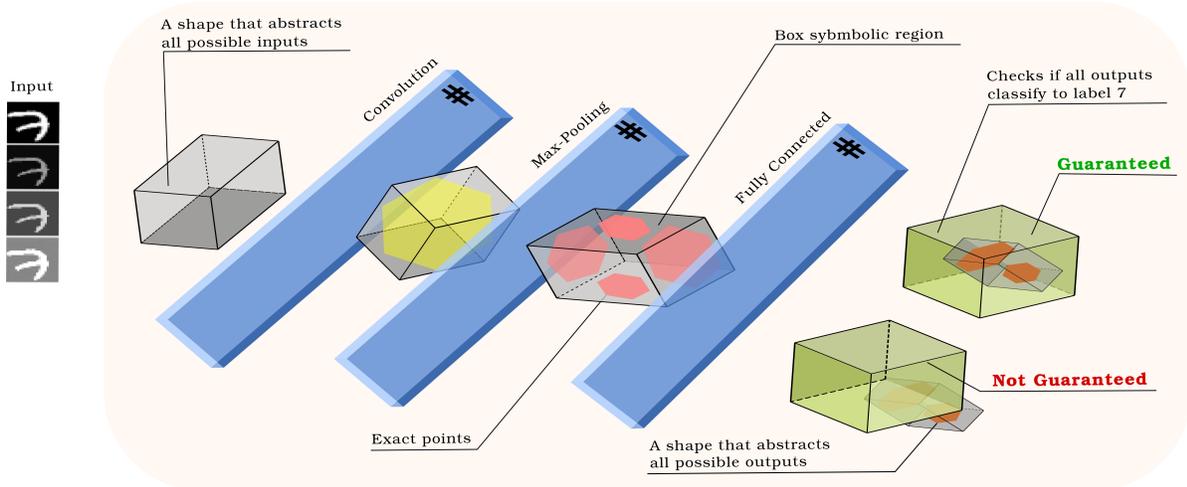}
  \caption{Illustration of abstract certification to verify all possible perturbations. Initially, the input distribution is approximated by an box abstract region. It is then propagated through layers through suitable box transformations. $\#$ shows that the transformations are abstract. At the last stage, it verify if all points in abstract region corresponds to the same output label.}
  \label{fig:abscert}
\end{figure*}
This certification is carried out using state-of-the-art sound, precise and scalable toolbox: \textbf{E}TH's \textbf{R}obustness \textbf{AN}alyzer (ERAN) \cite{singh2019eth}, which is a robustness analyzer based on abstract interpretation for verification of neural network model. 
The elements of the abstract certification are explained below: 
\begin{itemize}
\itemsep0em
    \item \textbf{Region Definition: } This is the most important decision in an abstract certification. The type of symbolic region determines the precision of the analysis. In practice, we aim for regions that are scalable to large networks and, at the same time, approximates the perturbation in the best possible way. The symbolic region in figure \ref{fig:abscert} is Box. Some examples of other regions are DeepZono and DeepPoly, which are inspired by the geometry of regular zonotope and polytope, respectively, and are modified in the context of neural networks. The DeepPoly is the most precise region but has expensive scalability. \\
    \textbf{Scalability: } Box $>$ DeepZono $>$ DeepPoly \\
    \textbf{Precision: } DeepPoly $>$ DeepZono $>$ Box\\
    The trade-off between scalability and precision exists, as definition complexity increase with precision leading to expensive time complexity. 
    \item \textbf{Region Transformation: } To work with symbolic regions, we define abstract operations for the transformations of these regions in a typical DNN. Some examples of transformations include affine, ReLU, Max-pool, sigmoid, etc. In Figure \ref{fig:abscert}, we observe the abstract transformations of box region. These transformations may lose precision during the operation, and hence, it is necessary to design the transformations carefully. We aim for exactness(completeness) and soundness in this process. However, most often, the scalable regions lose precision and hence are incomplete methods. As we see in Figure \ref{fig:abscert}, the colored regions inside the boxes are the actual region of the presence of data points; however, the box transformations lead to loss of precision at each step.
    \item \textbf{Verification: } The goal of the final stage is verification. In general, to have a guarantee, the output shape produced after all the transformations should satisfy the condition in \ref{eq:verify}:
    \begin{equation}
    \forall \; \emph{i} \in I, \emph{i} \vDash \phi \implies \mathcal{N}(\emph{i}) \vDash \psi
    \label{eq:verify}
    \end{equation}
    \noindent where $\emph{I}$ is the input space, and $\mathcal{N}(\emph{i})$ is the output after passing an input through neural network $\mathcal{N}$ transformations. In other words, every concrete point inside the output region should classify to the same label, which means the entire region has to lie inside the space represented by postcondition $\psi$ as shown in figure \ref{fig:abscert}.
\end{itemize}
The authors in \cite{gehr2018ai2}, \cite{singh2019abstract}, \cite{singh2018fast} and \cite{singh2019beyond} have shown detailed analysis of box, zonotope and polytope regions with their abstract transformations. Designing symbolic regions is an active area of research to achieve higher precision.

\section{Training Methodology}

\begin{figure}[h!]
	\hspace{-0.2cm}
	\scalebox{1}{
		\centering
\vspace*{-4mm}
\pgfdeclarelayer{background}
\pgfdeclarelayer{foreground}
\pgfsetlayers{background,main,foreground}
\tikzstyle{block} = [rectangle, draw, fill=blue!20, text centered, rounded corners, minimum height=4em,minimum width=8em,text width=5em]
\tikzstyle{bck} = [rectangle, draw, fill=blue!20, text centered, rounded corners, minimum height=4em,minimum width=8em,text width=6em]
\tikzstyle{bck1} = [rectangle, draw, fill=blue!20, text centered, rounded corners, minimum height=4em,minimum width=12em,text width=10em]
\tikzstyle{dia} = [diamond, draw, fill=blue!20, text centered, rounded corners, minimum height=2em,minimum width=4.5em,text width=4em]
\tikzstyle{diamon} = [diamond, draw, fill=blue!20, text centered, rounded corners, minimum height=3em,minimum width=6em,text width=5em]
\tikzstyle{ov} = [ellipse, draw, fill=blue!20, text centered, rounded corners, minimum height=4em,minimum width=7em,text width=4em]	
\tikzstyle{line} = [draw, -latex']
\begin{tikzpicture}[node distance = 3cm, auto,scale=0.5, every node/.style={transform shape}]
\node [ov,align=center,node distance = 2cm] (model) {Model};
\node [ov,right of=model,node distance = 5cm] (OD) {Original Data};
\node [ov,below of=model,node distance = 1.5cm, xshift=2.5cm] (property) {Define Property};
\node [bck1,below of=property, node distance = 2.5cm, align=center] (SA) {Symbolic Abstraction and Transformations};
\node [diamon, below of=SA,node distance = 2.9cm, align=center] (correct) {Correct?};
\node [diamon, below of=correct,node distance = 3.6cm, align=center] (verify) {Verified?};
\node [bck,right of=verify, node distance = 3.6cm, yshift=-4cm, align=center] (attack) {Perform Attack};
\node [bck,below of=verify, node distance = 5.5cm, align=center] (CDS) {Custom Dataset};
\node [bck,below of=CDS, node distance = 2.3cm, align=center] (TM) {Train Model};
\node [ov,below of = TM, align=center,node distance = 2.3cm] (val) {Validate};
\draw [-triangle 45] (model) |- (property);
\draw [-triangle 45] (OD) |- (property);
\draw [-triangle 45] (property) -- (SA);
\draw [-triangle 45] (SA) -- (correct);
\draw [-triangle 45] (correct) -- node[near start]{Yes}(verify);
\draw [-triangle 45] (correct) -| node[near start]{No}(attack);
\draw [-triangle 45] (CDS) -- (TM);
\draw [-triangle 45] (TM) -- (val);
\draw [-triangle 45] (verify) -|node[near start]{No} (attack);
\draw [-triangle 45] (verify) --node{Yes} (CDS);
\draw [-triangle 45] (attack) |- (CDS);
\begin{pgfonlayer}{background}
\path (model.west |- model.north)+(-1.7,1.1) node (a) {};
\path (OD.east |- verify.south)+(2.3,-0.9) node (b) {};
\path[fill=green!10,rounded corners, draw=black!50, dashed](a) rectangle (b); 
\node [above of=OD, node distance=0.9cm,yshift=0.5cm,xshift=-5cm,align=center] (d2) {\large\bf ABSTRACT CERTIFICATION}; 
\end{pgfonlayer} 
\begin{pgfonlayer}{background}
\path (SA.west |- SA.north)+(-2.5,0.6) node (a) {};
\path (SA.east |- verify.south)+(3.1,-0.6) node (b) {};
\path[fill=yellow!10,rounded corners, draw=black!50, dashed](a) rectangle (b); 
\node [above of=SA, node distance=0.9cm,yshift=0.0cm,xshift=-3.8cm,align=center] (d2) {\large\bf ERAN}; 
\end{pgfonlayer}
\begin{pgfonlayer}{background}
\path (model.west |- attack.north)+(-1.7,1) node (a) {};
\path (OD.east |- val.south)+(2.3,-0.5) node (b) {};
\path[fill=lightgray!10,rounded corners, draw=black!50, dashed](a) rectangle (b); 
\node [left of=CDS, node distance=3.7cm,yshift=2.5cm,xshift=-0.0cm,align=center] (d2) {\large \bf ADVERSARIAL\\
\large \bf TRAINING}; 
\end{pgfonlayer} 

\end{tikzpicture}
	}
	\caption{Our training procedure.}
	\label{fig:flow}
\end{figure}
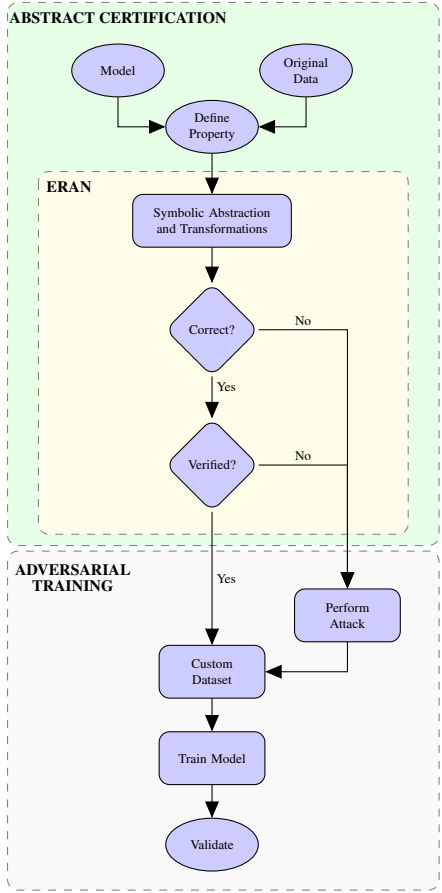

In this section, we describe the training strategy. As stated earlier, the goal is to use the subset of the original dataset for adversarial training. The following steps are involved in the proposed methodology:
\begin{itemize}
\itemsep0em
    \item Select a model of choice and required dataset
    \item Define a pre-condition $\phi$ bounded by $\epsilon$. The larger the value of $\epsilon$, the higher the risk of an attack
    \item Run the ERAN toolbox with the model, defined pre-condition, required symbolic region, and the whole training dataset 
    \item For every input in the training dataset, the toolbox initially checks whether the input is correctly classified or not. 
    \item Once it is correctly classified, the toolbox verifies the image. Hence, the dataset is partitioned into three subsets: incorrectly classified, correctly classified but unverified, and correctly classified, which are verified.
    \item In each training epoch, the inputs that were verified and classified correctly are not touched and are used for training using the natural method. 
    \item In the same training epoch, we perform training with adversarial examples from the group of inputs that are either incorrectly classified or correctly classified but unverified. 
    \item The model is finally validated using the test data
\end{itemize}
Adopting the training methodology ensures that that the verified samples are not attacked. Unlike conventional adversarial training, we do not disturb the whole input distribution. Hence, we have comparable robust performance, i.e., adversarial accuracy, to the conventional adversarial training using this methodology. However, the beauty of this methodology lies in the lesser sacrifice of natural accuracy. Figure \ref{fig:flow} shows the flow of the proposed training procedure.

\section{Experiments}
\label{sec:four}

In order to demonstrate our reasoning, we propose the following datasets, neural network architectures, and other conditions (attack type, symbolic region, etc.), establishing the environment required for the experiments.
The proposed training method is evaluated using the two datasets
\begin{itemize}
\itemsep0em
    \item \textbf{MNIST:} dataset for digit recognition \cite{lecun1998mnist} . The dataset contains 60,000 grayscale training images with 28 $\times$ 28 pixels. Each image is labelled as one of the ten digits (0-9). 
    \item \textbf{CIFAR10:} dataset of tiny images \cite{krizhevsky2009learning}. The data set set contains 5 batches of 10,000 training images. Every image consists of RGB channels each having 32 $\times$ 32 pixels, where each image belongs to one of the 10 different classes  (e.g, dog, cat, truck, ship etc.)
\end{itemize}

\noindent We have considered a small, simple, and similar convolution-based architecture for both MNIST and CIFAR10 datasets for neural network models. The simplicity of the data sets will ensure less time to be spent on training and will help to focus more on the method. However, the idea is general and can be easily implemented for other data sets of interest. In order to define the convolutional layer, we use a notation $N_{p \times q}$, where N is the filters present in each layer and p and q being the dimension of the kernel. Similarly, for a fully connected layer, we use another notation $A \times B$, where A is the number of layers present in the network and B is the total neurons in each layer. We are interested in defining our architectures in terms of neurons present in them because the time complexity of certification largely depends on the neuron present in the model. The proposed architecture for our preliminary study is given in table \ref{tab:arch}.
For training, adversarial examples are generated by attacking the inputs samples. In our case, we have used a PyTorch based attack library named \textit{Torchattacks} \cite{kim2020torchattacks}. It consists of many options to select an attack. However, we have settled for the old famous PGD attack in our study. The results refer to conventional adversarial training as 'Adversarial Training' and our methodology as 'Our Training.' We are limiting ourselves to DeepPoly abstractions for symbolic regions due to their high precision compared to other symbolic regions. 
\begin{table}[h]
\begin{small}
\begin{center}
\caption{Model Architecture}\label{tab:arch}
  \renewcommand{\arraystretch}{1}
 \renewcommand\normalsize{\small}
\begin{tabular}{|l|l|l|l|}
    \hline
      \multicolumn{2}{|c|}{MNIST} &
      \multicolumn{2}{|c|}{CIFAR10} \\
    \hline
    Shape & Neurons &Shape & Neurons \\
    \hline
    $16_{4\times4 }$ &  & $16_{4\times4 }$ &  \\
    $32_{4\times4 }$ &  & $32_{4\times4 }$ &  \\
    $1\times800$ & 3,614 & $1\times1152$ & 4,862 \\
    $1\times100$ &  & $1\times100$ &  \\
    $1\times10$ &  & $1\times10$ &  \\
    \hline
    
  \end{tabular}
 \end{center}
\end{small}
\end{table}
\begin{figure}[h]
  \centering
  \includegraphics[width=1\linewidth]{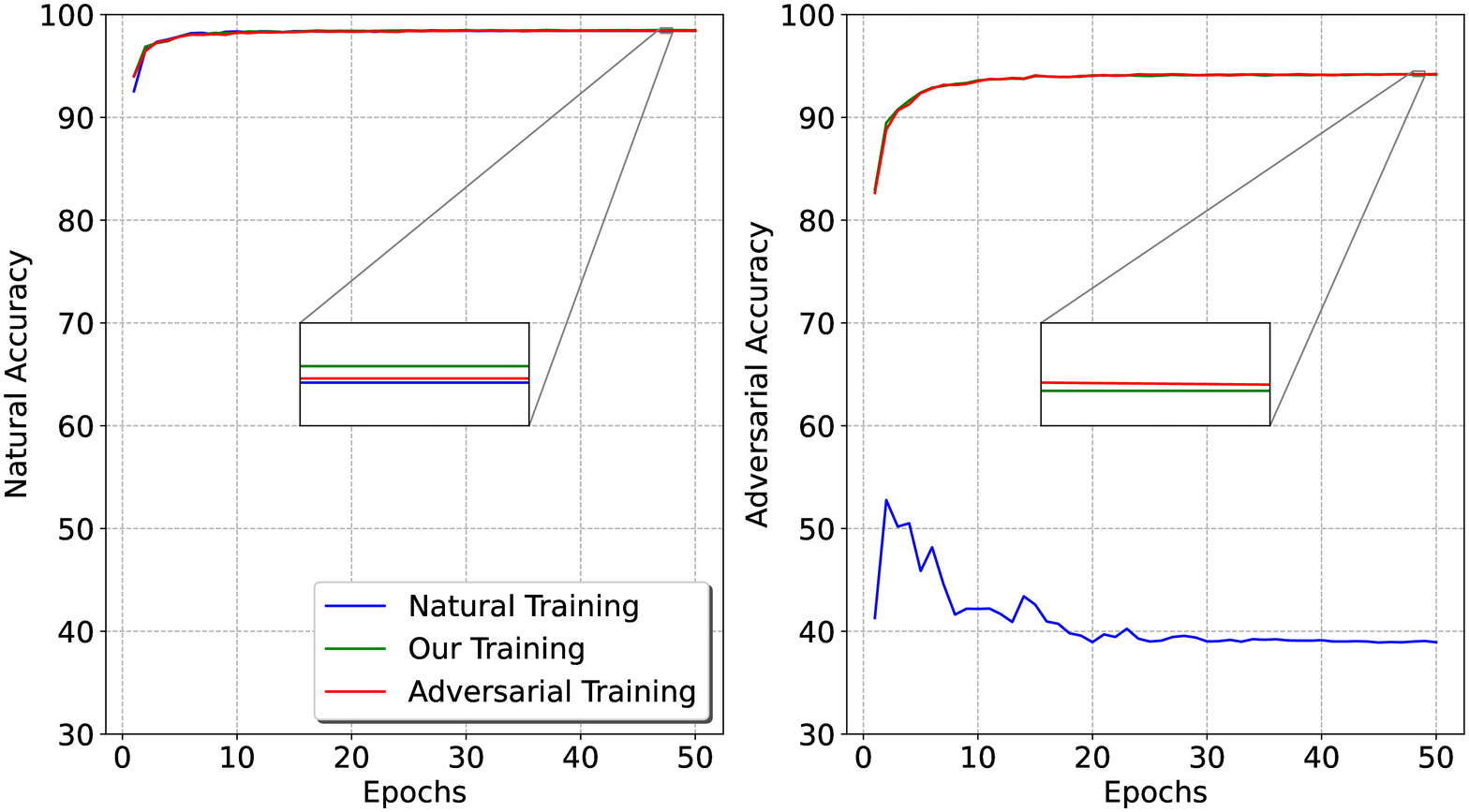}
  \caption{ MNIST Trial: Epochs:50 $\epsilon$:0.1 Steps:20 $\alpha:0.01$ }
  \label{fig:remnist}
\end{figure}
\begin{figure*}[t]
\begin{subfigure}{.5\textwidth}
  \centering
  \includegraphics[width=1\linewidth]{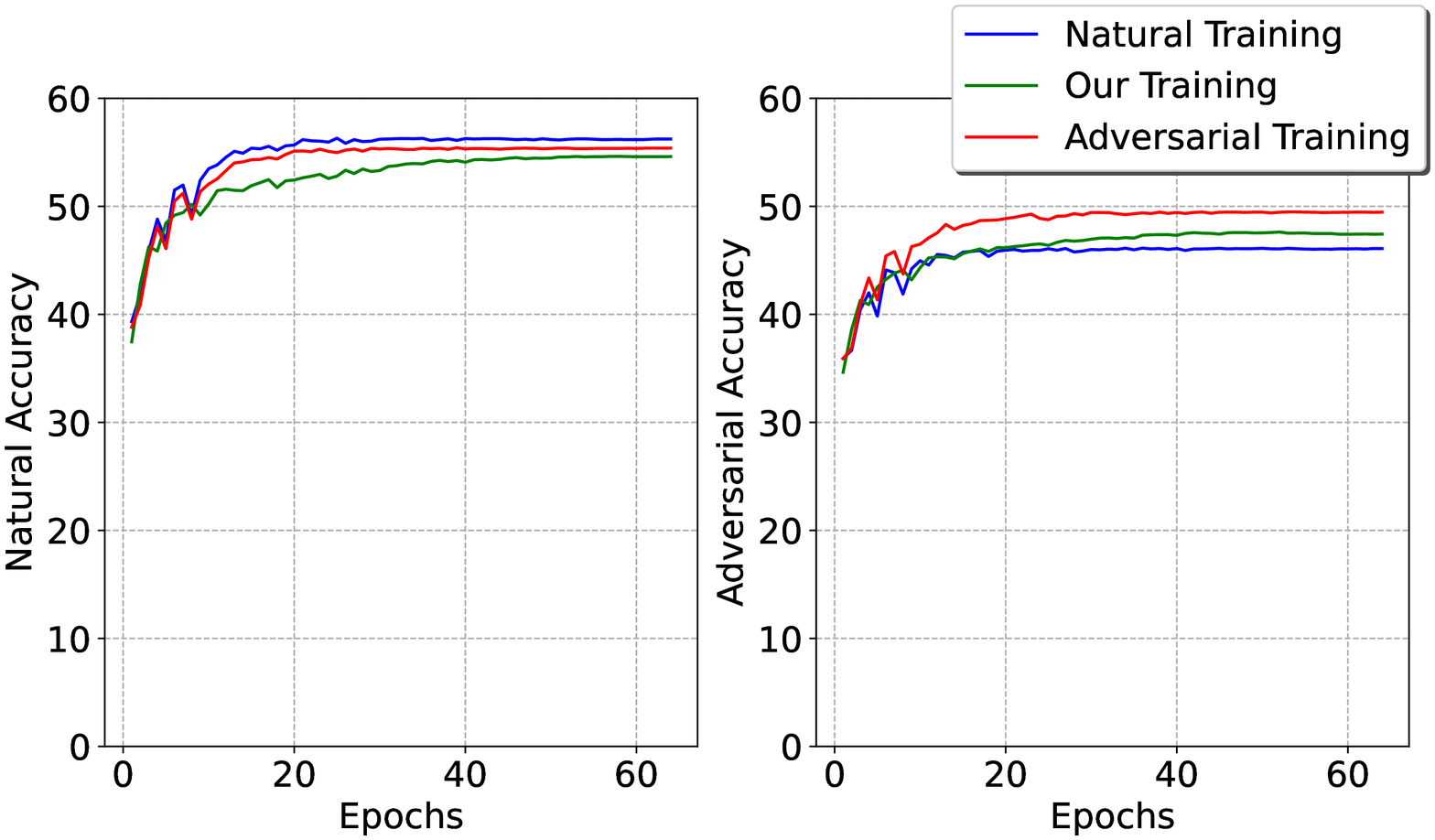}  
  \caption{Epochs:64 $\epsilon$:1/255 Steps:15 $\alpha:0.001$}
  \label{fig:sub-first}
\end{subfigure}
\begin{subfigure}{.5\textwidth}
  \centering
  \includegraphics[width=1\linewidth]{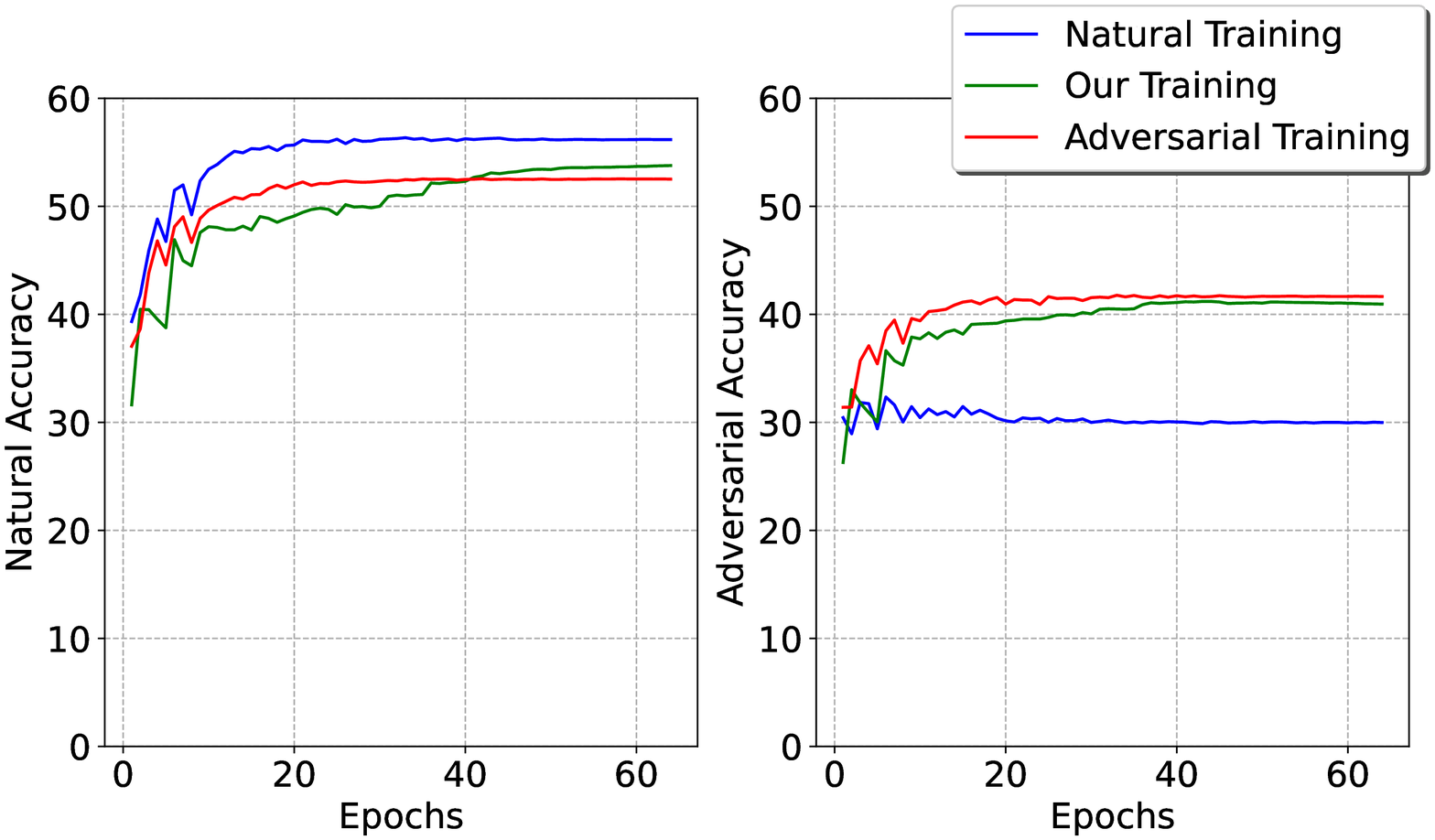}  
  \caption{Epochs:64 $\epsilon$:2/255 Steps:15 $\alpha:0.001$}
  \label{fig:sub-second}
\end{subfigure}
\vspace{0.2cm}
\begin{subfigure}{.5\textwidth}
  \centering
  \includegraphics[width=1\linewidth]{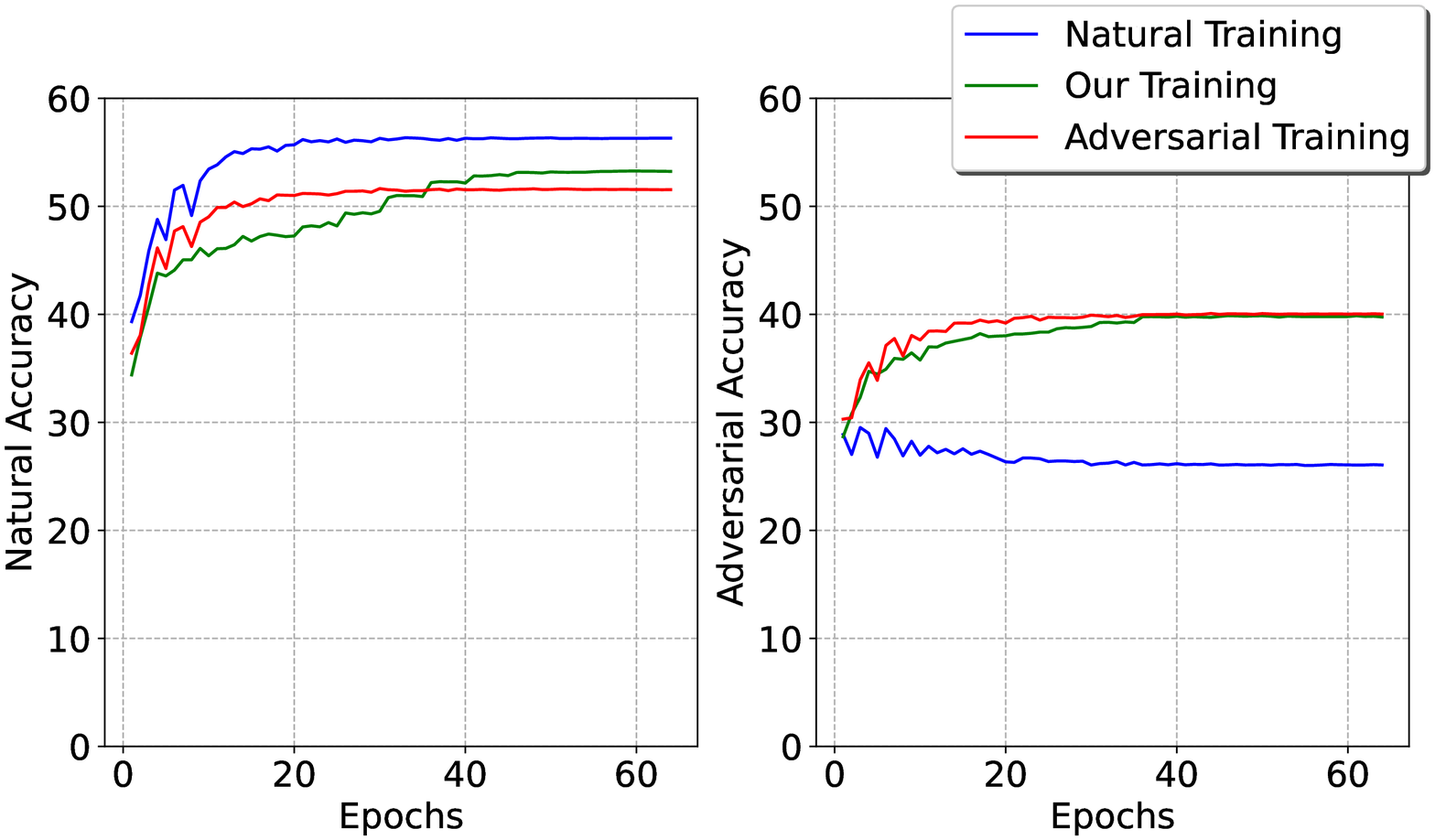}  
  \caption{Epochs:64 $\epsilon$:4/255 Steps:15 $\alpha:0.005$}
  \label{fig:sub-third}
\end{subfigure}
\begin{subfigure}{.5\textwidth}
  \centering
  \includegraphics[width=1\linewidth]{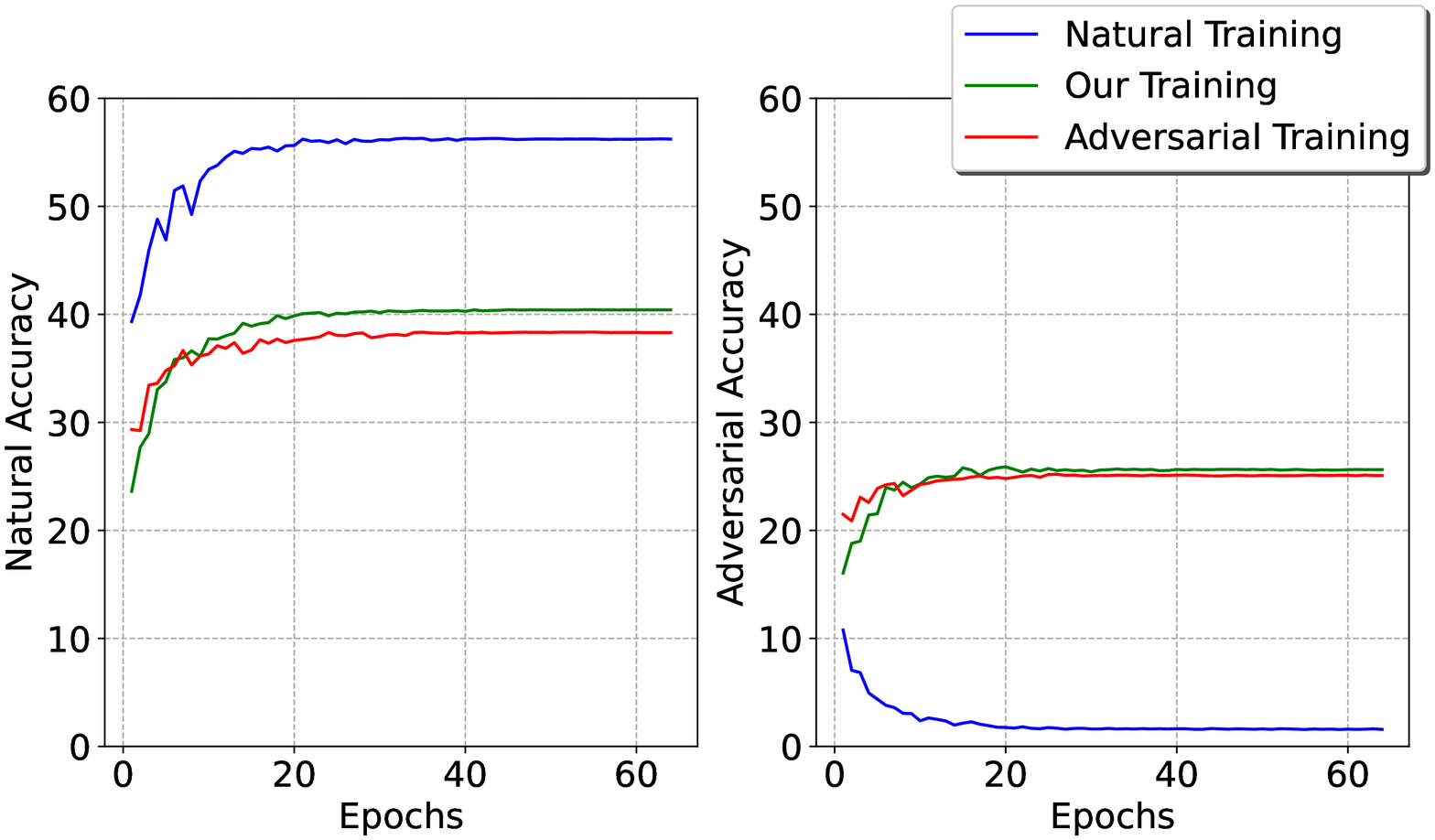}  
  \caption{Epochs:64 $\epsilon$:8/255 Steps:15 $\alpha:0.005$}
  \label{fig:sub-fourth}
\end{subfigure}
\vspace{0.1cm}
\caption{CIFAR10 Trial}
\label{fig:fig}
\end{figure*}
Additionally, the toolbox has a variant of DeepPoly called GPUPoly for fast verification of models on GPU.
For MNIST, we have conducted the trial with 20 iterations of PGD attack with a step size of 0.01. The adversarial training and evaluation are carried out with $l_{\infty}$ bound ($\epsilon$) of 0.1 for 50 epochs. Figure \ref{fig:remnist} shows the result of the trial. We observe that for the MNIST dataset, the improvement is not considerable. Also, when we tried to increase $\epsilon$ beyond 0.1, the subset of verified inputs was minimal to produce considerable improvement against conventional adversarial training. Here we observed a reasonable trade-off where increasing $\epsilon$ lead to meager verified samples while reducing $\epsilon$ led to the insignificant difference between natural and adversarial accuracy.\\
\indent For CIFAR 10, we conducted the experiments with 15 iterations of the PGD attack. The adversarial training and evaluation is carried out with $l_{\infty}$ bounds ($\epsilon$) of 1/255 (Figure \ref{fig:sub-first}), 2/255 (Figure \ref{fig:sub-second}), 4/255 (Figure \ref{fig:sub-third}), 8/255 (Figure \ref{fig:sub-fourth}) respectively for 64 epochs. We observe that the accuracy achieved is less as we have used a primitive and small convolutional model. However, our experiments aim to demonstrate the boost of natural accuracy. Although not very significant, we have observed around 1.5$\%$ - 2$\%$ improvement in Figure \ref{fig:sub-third} and Figure \ref{fig:sub-fourth}. Since the accuracy achieved is lower, the verified samples being a subset tends to be lower. Hence, using powerful models will improve accuracy, leading to a larger set of verified inputs. It naturally increases the possibility of achieving significant improvement in natural accuracy. However, unsurprisingly, the need for computational resources and the training time also increases, and this analysis will be a necessary component of our future work. We also observe the similar trade-off as observed with MNIST dataset; the number of verified samples becomes less as we increase $\epsilon$. Specifically, when $\epsilon$ is made to 10/255, we observed no verified samples through abstract certification leading to the comparable performance by 'Our Training' and 'Adversarial Training' in terms of natural and adversarial accuracy.
\section{Conclusions}
\label{sec:five}
\label{sec:conclusion}
In this position paper, we have demonstrated the utilization of abstract certification for training neural networks against adversarial attacks. The proposed framework is practical and straightforward to boost the natural accuracy in applications where natural accuracy is as important as the defense against adversaries. The preliminary results pave the way for the broader research that can be done in this area. The study aimed to initially evaluate the feasibility of the framework implementation and then understand the scope of improvement in the performance. We have achieved satisfactory results with the initial set of experiments; however, we aim to validate this work in future with more concrete results. Hence, we propose some of the possible directions to extend our work:
\begin{itemize}
\itemsep0em
    \item Train with powerful models to observe the considerable improvement in accuracy
    \item Train with different types and variants of attacks to understand the generalization of the method.  Also, training defense against geometric adversaries will be interesting as the ERAN toolbox supports geometric certification.
    \item Extending the work to real-life datasets will give better insights into this framework. Even the simplest of the models already have a good performance on simple datasets like MNIST. 
    \item Formulate hyper-parameter tuning to get right $\epsilon$ for abstract certification to achieve maximum improvement in accuracy
\end{itemize}

\section{Acknowledgement}

We gratefully acknowledge the support from the advanced research computing platform: Sockeye at The University of British Columbia for the resource allocation for this study.

{\small
\bibliographystyle{ieee_fullname}
\bibliography{main}

\begin{thebibliography}{10}\itemsep=-1pt

\bibitem{akhtar2018threat}
Naveed Akhtar and Ajmal Mian.
\newblock Threat of adversarial attacks on deep learning in computer vision: A
  survey.
\newblock {\em Ieee Access}, 6:14410--14430, 2018.

\bibitem{brosnan2004improving}
Tadhg Brosnan and Da-Wen Sun.
\newblock Improving quality inspection of food products by computer vision----a
  review.
\newblock {\em Journal of food engineering}, 61(1):3--16, 2004.

\bibitem{cousot1977abstract}
Patrick Cousot and Radhia Cousot.
\newblock Abstract interpretation: a unified lattice model for static analysis
  of programs by construction or approximation of fixpoints.
\newblock In {\em Proceedings of the 4th ACM SIGACT-SIGPLAN symposium on
  Principles of programming languages}, pages 238--252, 1977.

\bibitem{dong2018feature}
Guozhu Dong and Huan Liu.
\newblock {\em Feature engineering for machine learning and data analytics}.
\newblock CRC Press, 2018.

\bibitem{gehr2018ai2}
Timon Gehr, Matthew Mirman, Dana Drachsler-Cohen, Petar Tsankov, Swarat
  Chaudhuri, and Martin Vechev.
\newblock Ai2: Safety and robustness certification of neural networks with
  abstract interpretation.
\newblock In {\em 2018 IEEE Symposium on Security and Privacy (SP)}, pages
  3--18. IEEE, 2018.

\bibitem{goodfellow2014explaining}
Ian~J Goodfellow, Jonathon Shlens, and Christian Szegedy.
\newblock Explaining and harnessing adversarial examples.
\newblock {\em arXiv preprint arXiv:1412.6572}, 2014.

\bibitem{kamilaris2018deep}
Andreas Kamilaris and Francesc~X Prenafeta-Bold{\'u}.
\newblock Deep learning in agriculture: A survey.
\newblock {\em Computers and electronics in agriculture}, 147:70--90, 2018.

\bibitem{khan2018guide}
Salman Khan, Hossein Rahmani, Syed Afaq~Ali Shah, and Mohammed Bennamoun.
\newblock A guide to convolutional neural networks for computer vision.
\newblock {\em Synthesis Lectures on Computer Vision}, 8(1):1--207, 2018.

\bibitem{kim2020torchattacks}
Hoki Kim.
\newblock Torchattacks: A pytorch repository for adversarial attacks.
\newblock {\em arXiv preprint arXiv:2010.01950}, 2020.

\bibitem{krizhevsky2009learning}
Alex Krizhevsky, Geoffrey Hinton, et~al.
\newblock Learning multiple layers of features from tiny images.
\newblock 2009.

\bibitem{lecun1998mnist}
Yann LeCun.
\newblock The mnist database of handwritten digits.
\newblock {\em http://yann. lecun. com/exdb/mnist/}, 1998.

\bibitem{madry2017towards}
Aleksander Madry, Aleksandar Makelov, Ludwig Schmidt, Dimitris Tsipras, and
  Adrian Vladu.
\newblock Towards deep learning models resistant to adversarial attacks.
\newblock {\em arXiv preprint arXiv:1706.06083}, 2017.

\bibitem{nguyen2015deep}
Anh Nguyen, Jason Yosinski, and Jeff Clune.
\newblock Deep neural networks are easily fooled: High confidence predictions
  for unrecognizable images.
\newblock In {\em Proceedings of the IEEE conference on computer vision and
  pattern recognition}, pages 427--436, 2015.

\bibitem{otter2020survey}
Daniel~W Otter, Julian~R Medina, and Jugal~K Kalita.
\newblock A survey of the usages of deep learning for natural language
  processing.
\newblock {\em IEEE Transactions on Neural Networks and Learning Systems},
  32(2):604--624, 2020.

\bibitem{papernot2016transferability}
Nicolas Papernot, Patrick McDaniel, and Ian Goodfellow.
\newblock Transferability in machine learning: from phenomena to black-box
  attacks using adversarial samples.
\newblock {\em arXiv preprint arXiv:1605.07277}, 2016.

\bibitem{shafahi2019adversarial}
Ali Shafahi, Mahyar Najibi, Amin Ghiasi, Zheng Xu, John Dickerson, Christoph
  Studer, Larry~S Davis, Gavin Taylor, and Tom Goldstein.
\newblock Adversarial training for free!
\newblock {\em arXiv preprint arXiv:1904.12843}, 2019.

\bibitem{singh2019beyond}
Gagandeep Singh, Rupanshu Ganvir, Markus P{\"u}schel, and Martin Vechev.
\newblock Beyond the single neuron convex barrier for neural network
  certification.
\newblock 2019.

\bibitem{singh2018fast}
Gagandeep Singh, Timon Gehr, Matthew Mirman, Markus P{\"u}schel, and Martin~T
  Vechev.
\newblock Fast and effective robustness certification.
\newblock {\em NeurIPS}, 1(4):6, 2018.

\bibitem{singh2019abstract}
Gagandeep Singh, Timon Gehr, Markus P{\"u}schel, and Martin Vechev.
\newblock An abstract domain for certifying neural networks.
\newblock {\em Proceedings of the ACM on Programming Languages}, 3(POPL):1--30,
  2019.

\bibitem{singh2019eth}
Gagandeep Singh, M Mirman, T Gehr, A Hoffman, P Tsankov, D Drachsler-Cohen, M
  P{\"u}schel, and M Vechev.
\newblock Eth robustness analyzer for neural networks (eran), 2019.

\bibitem{voulodimos2018deep}
Athanasios Voulodimos, Nikolaos Doulamis, Anastasios Doulamis, and Eftychios
  Protopapadakis.
\newblock Deep learning for computer vision: A brief review.
\newblock {\em Computational intelligence and neuroscience}, 2018, 2018.

\bibitem{wong2020fast}
Eric Wong, Leslie Rice, and J~Zico Kolter.
\newblock Fast is better than free: Revisiting adversarial training.
\newblock {\em arXiv preprint arXiv:2001.03994}, 2020.

\bibitem{xu2020adversarial}
Han Xu, Yao Ma, Hao-Chen Liu, Debayan Deb, Hui Liu, Ji-Liang Tang, and Anil~K
  Jain.
\newblock Adversarial attacks and defenses in images, graphs and text: A
  review.
\newblock {\em International Journal of Automation and Computing},
  17(2):151--178, 2020.

\bibitem{you2016image}
Quanzeng You, Hailin Jin, Zhaowen Wang, Chen Fang, and Jiebo Luo.
\newblock Image captioning with semantic attention.
\newblock In {\em Proceedings of the IEEE conference on computer vision and
  pattern recognition}, pages 4651--4659, 2016.

\end{thebibliography}
}

\end{document}